\documentclass[letterpaper, 10 pt, conference]{ieeeconf}  
\IEEEoverridecommandlockouts     
\usepackage{import}
\usepackage{cite}
\usepackage{graphics} 
\usepackage{epsfig} 
\usepackage{times} 
\usepackage{amsmath} 
\usepackage{amssymb}  
\usepackage{lipsum}
\usepackage{multicol}
\usepackage{graphicx}
\usepackage{afterpage}
\usepackage{subcaption}
\usepackage{multirow}
\usepackage{layout}
\usepackage{gensymb}
\usepackage{graphicx}
\usepackage[dvipsnames]{xcolor}
\usepackage{amsmath}
\usepackage{varwidth}
\usepackage{tikz}
\usetikzlibrary{shapes,arrows,positioning}
\usepackage{booktabs}

\usepackage[utf8]{inputenc} 
\usepackage[ngerman]{babel} 
\usepackage[T1]{fontenc}    

\usepackage{siunitx}
\usepackage[nolist]{acronym}
\usepackage{hyperref}

\usepackage{fancyhdr}
\fancypagestyle{pageStyleOne}{%
    \fancyhf{}
}


\begin{document}
\vspace*{0.35in}

\begin{flushleft}
{\Large
\textbf\newline{BioTracker: An Open-Source Computer Vision Framework for Visual Animal Tracking}
}
\newline
\\
Hauke J. Mönck\textsuperscript{1},
Andreas Jörg\textsuperscript{1,2},
Tobias von Falkenhausen\textsuperscript{1},
Julian Tanke\textsuperscript{1},
Benjamin Wild\textsuperscript{1},
David Dormagen\textsuperscript{1},
Jonas Piotrowski\textsuperscript{1},
Claudia Winklmayr\textsuperscript{1,3}
David Bierbach\textsuperscript{4},
Tim Landgraf\textsuperscript{1,\*},
\\
\bigskip
\bf{1} Freie Universität Berlin, FB Mathematik u. Informatik, Arnimallee 7, 14195 Berlin, Germany
\\
\bf{2} University of Applied Science Kempten, Bahnhofstraße 61, D-87435, Germany
\\
\bf{3} Humboldt-Universität zu Berlin, Bernstein Center for Computational Neuroscience, Unter den Linden 6, 10099 Berlin
\\
\bf{4} Leibniz-Institute of Freshwater Ecology and Inland Fisheries, Müggelseedamm 310, 12587 Berlin, Germany
\\
\bigskip
*corresponding author: tim.landgraf@fu-berlin.de
\\
\bigskip
Keywords: computer vision, tracking software, framework, animal tracking, animal behavior, ecology, guppy, molly, zebrafish
\bigskip

\end{flushleft}

\section*{Abstract}
The study of animal behavior increasingly relies on (semi-) automatic methods for the extraction of relevant behavioral features from video or picture data. To date, several specialized software products exist to detect and track animals' positions in simple (laboratory) environments. Tracking animals in their natural environments, however, often requires substantial customization of the image processing algorithms to the problem-specific image characteristics. Here we introduce BioTracker, an open-source computer vision framework, that provides programmers with core functionalities that are essential parts of a tracking software, such as video I/O, graphics overlays and mouse and keyboard interfaces. BioTracker additionally provides a number of different tracking algorithms suitable for a variety of image recording conditions. The main feature of BioTracker is however the straightforward implementation of new problem-specific tracking modules and vision algorithms that can build upon BioTracker's core functionalities. With this open-source framework the scientific community can accelerate their research and focus on the development of new vision algorithms.

\section*{Introducing BioTracker}
Animal tracking, e.g., extracting the spatial position or even specific movement patterns from a video source, has become increasingly popular in biological research \cite{dell_automated_2014}, which is reflected in an increasing number of software solutions available. Many of these software programs can handle only a specific tracking problem and are tailored to the image processing tasks defined by the experimental conditions and the animal models under observation. 

Throughout the last decade we encountered several image processing tasks that could not be handled by existing tracking algorithms: cascades of ripples on water, fish shoals in the wild, biomimetic robots, bumblebees or honey bees in their colonies, either with or without markers.\cite{landgraf_tracking_2007} \cite{hussaini_sleep_2009} \cite{jin_walking_2014} \cite{landgraf_blending_2014} \cite{wario_automatic_2015} \cite{landgraf_robofish:_2016} \cite{wario_automatic_2017} \cite{bierbach_insights_2018} \cite{wild_automatic_2018} While each tracking task was solved with a custom algorithm, a significant portion of the source code comprised functionalities that were required in all applications, such as the user interface, interaction handling, reading and writing video data and exporting tracking results. Developing new tracking solutions therefore involved reimplementing the same components, effectively consuming development resources for the actual image processing algorithm.
\\
We therefore developed the open-source computer vision framework “BioTracker” that implements several functionalities for reoccurring tasks and allows to dynamically load specific tracking algorithms that can call framework components for displaying results or user interactions. 

Such a separation of core components from dynamically loaded modules is not only beneficial for developers but also further helps users to find the best suitable tracking solution for their problems at hand. With BioTracker, testing several existing algorithms (or even developing new ones) for a given use-case does not require installing new applications or learning new interaction procedures. Users with several different use-cases thus benefit from using the same look-and-feel and potentially save time customizing their subsequent analyses to different tracking output formats.

A similar approach is taken by “SwisTrack” \cite{correll_swistrack:_2006} and “AnTracks” (www.antracks.org), which allow constructing a custom processing pipeline via a graphical user interface. Graphical programming might serve developers in exploring new algorithms, but does not provide the same flexibility as a native coding environment. Also, to be able to use these building blocks meaningfully, it may require some technical knowledge, which constrains the target audience.

The Java-based framework “ImageJ” \cite{schindelin_imagej_2015} provides a similar framework to the developer but is focused on processing single images rather than tracking objects in a sequence of images. It nonetheless does allow the implementation of tracking algorithms, but some of the core features, such as visualization and data export, have to be implemented on the module level.

The BioTracker has been employed successfully in some of the aforementioned tracking tasks. For instance, it has been used in the tracking of fish in an experimental tank as done in \cite{bierbach_insights_2018}. Here the BioTracker was part of a closed loop interaction scenario as a live tracking instance. Details on how to get and operate the program can be found here: \url{https://github.com/BioroboticsLab/biotracker_core/wiki}

\section*{Software Design}
BioTracker consists of four components.
\\
The “core” component provides the shared functionality, being the key to reducing implementation overhead. Functionality such as loading external trackers as plugins, reading videos, pictures and camera streams, and exporting tracking data to a generic format can be done. It also provides various standard graphical user interfaces such as playback, seeking, recording of videos, zooming images, and manipulating tracking objects. Multiple objects can be tracked while basic trajectory manipulations are possible. For instance adding, deleting, moving, correcting and annotating trackpoints can be done. 
\\
The “tracking module” component implements the respective tracking and image processing algorithm. This is done by implementing an interface loaded by the core component. 
The ”interfaces” component acts as a connector between core and tracking component by defining interaction methods between tracking module and core.
\\
The optional “utility” component provides functionality for the most common tracking issues to simplify tracking module implementation. 
\\
The core and interfaces depend on OpenCV and Qt as external libraries and the core additionally on Boost. Consequently tracking modules have to depend on Qt and OpenCV as well. The code conforms to C++11 and targets the largest build platforms Windows, Linux and Mac and follows the Model-View-Controller pattern.

\section*{Graphical User Interface}

\begin{figure*}
  \centering
    \includegraphics[width=\textwidth]{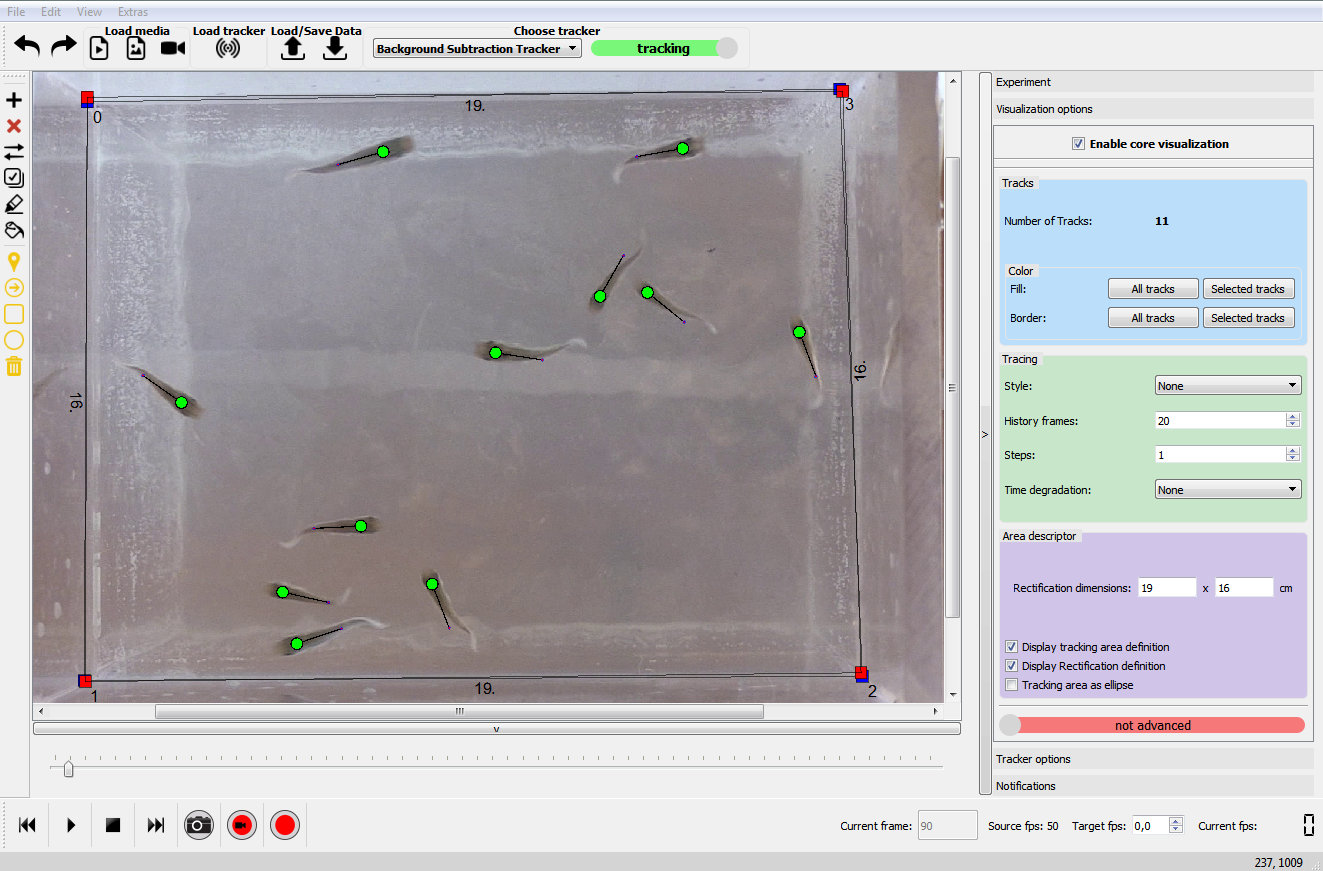}
  \caption[BioTracker]{Main view of the BioTracker. Here 11 sulfur mollies (Poecilia sulphuraria) are being tracked using the Background Subtraction Tracker. The green circles indicate location of a tracked entity, while the black line indicates orientation. Tracks and annotations handling can be done using the control bar to the left. Video playback is handled using the bottom control bar. The top control bar allows data handling and tracker selection. Finally on the right hand side visualization options are presented in the depicted tab, as well as tracking module specific parameters in another tab. }
  \label{fig:BT3}
\end{figure*}

The GUI is split into three parts. 

The first part is the tracking area which displays generic overlay and custom overlay on top of the input. The former contains generic trajectory visualization, as illustrated in figure \ref{fig:BT3}, handles mouse events and coordinate rectification. The latter is optionally provided by the tracking component to extend or replace these core visualizations. Mouse events cover e.g. right clicks on the tracking area, which will present a menu providing basic control over the tracking data and visualization. Entities can be added, removed, marked, have their color and transparency changed, etc.
\\
The second part are options, presented on the right hand side. This includes the view control, which modulates generic visualization in the tracking area. Other tabs are reserved for experiment control, notifications and parameters of the tracking component, e.g. binarization thresholds for background subtraction. 
\\
The last part are the control toolbars, which can be moved around freely. The video controls bar manages all the playback and recording aspects. 
The managing includes starting, stopping, seeking, recording the video with or without the overlay and more. It also displays the current frame number and playback speed and grants control over the maximum playback speed.
The content controls allow for instance adding tracked elements or annotations, loading of videos and trackers and saving tracked data. \\
We implemented a simple visualization for tracked objests as points with tails indicating movement direction for generic point-like data, such as illustrated in the scenario in figure \ref{fig:BT3}. More generic methods are to be implemented, e.g. for polygon shaped objects.

\subsection*{Input rectification and the coordinate system}
Tracking systems are usually required to produce world coordinates (cm or mm) rather than image coordinates (pixels). When recording a planar setup at an angle, the image of the arena is distorted. We have implemented a basic camera calibration procedure which is used to rectify this perspective distortion. Tracking modules can use these calibration and rectification features optionally and therefore handle different camera setups correctly. 

\subsection*{Output}
Detections can be saved for further analysis generically as comma-separated values (.csv), Json and a binary format, including an option to load and edit previous tracking results. This is done by serializing Qt properties one after another. This way developers can extend existing types and annotate which data needs to be stored additionally. It is also possible to record the video, optionally zoomed in or including overlaid tracking information. As BioTracker can handle live video streams from various USB cameras, it may also serve as a free-to-use video recorder.

\section*{Available tracking modules}
The BioTracker includes two different example trackers: \\ 
(1) The “Background Subtraction Tracker” implements a well-known algorithm used best for tracking tasks with contrastive objects on uniform background. This method covers standard scenarios, such as animals on a static, white background. It has been used successfully in different tracking software such as ToxTrack, \cite{rodriguez_toxtrac:_2017} and publications such as \cite{kaewtrakulpong_improved_2001}. Every image is integrated into the previous background according to a weight variable. For detection, new images are subtracted from the background, binarized and filtered. Then, ellipses are fitted to the blobs found in the foreground image via image moments \cite{bradski_gary_r._computer_1998}. 
The algorithm can be susceptible for sudden changes in lighting (such as large shadows or sudden spotlights) and movement of the camera. It works stably on slowly changing background (e.g. slow change in ambient lighting) and single, static objects on the background (e.g. dirt, maze structures). This tracker analyzes successive images separately and merges the detected ellipses from adjacent frames to a track.
\\ 
(2) The second example is the “Lucas-Kanade Tracker”. The algorithm computes image derivatives of designated points across time, as introduced by \cite{lucas_iterative_1981} according to \cite{bouguet_jean-yves_pyramidal_2000}.
Motion of an object is estimated via the optical flow in an region of interest around it. This is susceptible to fast movement, but does not require static background.

\section*{Evaluating the output}

\begin{figure}
  \centering
    \includegraphics[width=0.5\textwidth]{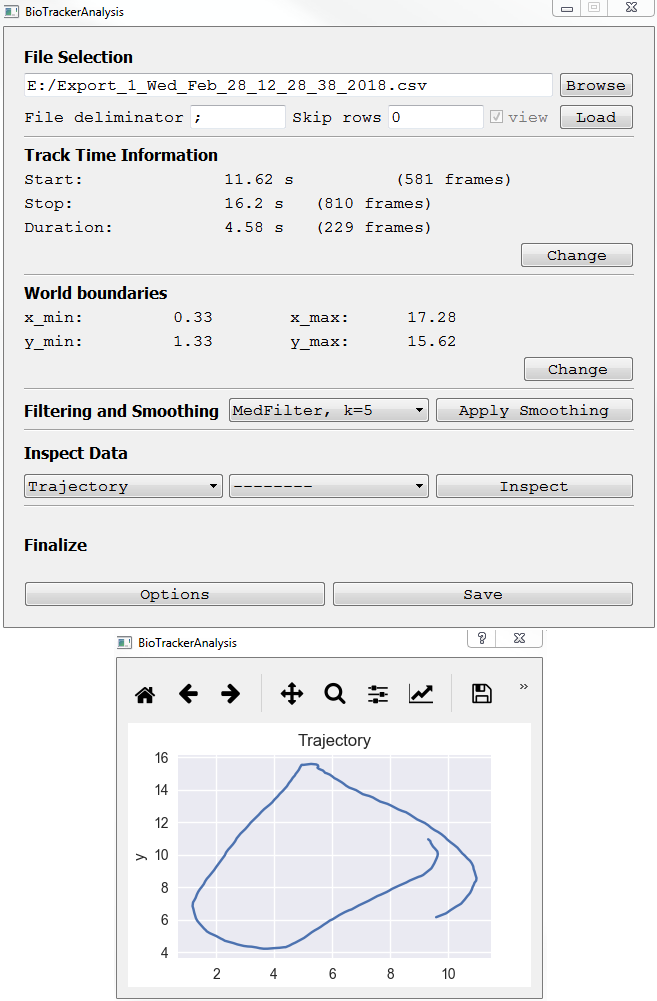}
  \caption[BioTracker]{The python based “Data Analysis Module” can be used to evaluate data created by the BioTracker and any other application producing .csv trajectories. The "File Selection" section allows loading files and defining agents and their respective data columns. The user can also select time and location slices to gain a more in-depth insight into the data. Applicable metrics can be plotted ad-hoc, as well as writing them to files along with .csv data including all selected metrics. }
  \label{fig:RoboStat}
\end{figure}

Along with the BioTracker comes a python tool named “Data Analysis Module” to evaluate the CSV output produced by the tracking modules. It has a simplistic user interface and provides an easy way to calculate a number of metrics for arbitrary CSV files by using user annotated columns. Metrics like speed or agent trajectories can be plotted ad-hoc.  
It calculates metrics like speed, inter-individual distance, transfer entropy and cross-correlation for all individuals. This is done pairwise where applicable and is written back to new CSV files. There is also the option to filter arbitrary columns and plot those metrics (see figure \ref{fig:RoboStat}). 
The Data Analysis Module stands alone and can handle and analyze CSV files from different sources, apart from the BioTracker.

\section*{Conclusion}
BioTracker is a novel framework to streamline implementation of tracking and image processing applications while allowing developers to focus on the task at hand - the tracking algorithms.
On one hand the software design is simplistic to serve minimalistic trackers. This has been showcased in the Lucas-Kanade tracker. On the other hand it it is rich in generic, optional features to accommodate for most unique and complex tasks. Algorithms profit from the performance of a native C++ implementation, which makes the BioTracker a feasible solution of real-time application as well as computation extensive tasks. 
The GUI implements common features for user interaction and encapsulates data serialization and management. As a crowd-improved open source platform the probability of introducing bugs in trivial features is low while granting a maximum of flexibility, standardized formats and usage patterns.
Two sample trackers showcase implementation and usage of the framework including a
third simplistic tracker for demo purposes. In order to enable users to go the whole way from having an animal on video to its tracked movement parameters, BioTracker also provides a tool to analyze and visually evaluate trajectories.

\section*{Acknowledgements}
The BioTracker as part of the RoboFish project was supported by the DFG (to TL: LA 3534/1-1, to DB: BI 1828/2-1). Furthermore, HM and DD have been supported by the Andrea von Braun foundation through a PhD fellowship. Special thanks goes to all the testers of the BioTracker software, as well as students implementing the foundations of the Background Subtraction Tracker.

\section*{Authors’ contributions}
\noindent
Conceived the idea: TL \\ 
Supervised the work flow: TL, HM, DB, BW \\ 
Software developed: HM, AJ, TF, JT, BW, DD, TL \\ 
Visualization: JP, TF, JT, AJ, TL \\  
Python evaluation (Data Analysis Module): CW \\ 
Wrote the manuscript: HM, DB, TL \\ 
Tested the software: BW, DD, DB, TL, JP \\  

\bibliography{library}

\bibliographystyle{apalike} 

\end{document}